\documentclass[runningheads]{llncs}
\pdfoutput=1
\usepackage{graphicx}
\usepackage{amsmath,amssymb} % define this before the line numbering.
\usepackage{subfig}
\usepackage{color}
\usepackage{hyperref}
\usepackage[mode=buildnew]{standalone}
\usepackage{array}
\usepackage[numbers]{natbib}
\usepackage[title,toc,titletoc,header]{appendix}
\usepackage{afterpage}
\usepackage{pdflscape}
\usepackage{tabularx}
\usepackage{rotating}

\newcommand*\samethanks[1][\value{footnote}]{\footnotemark[#1]}
\usepackage[width=122mm,left=12mm,paperwidth=146mm,height=193mm,top=12mm,paperheight=217mm]{geometry}
\begin{document}
% \linenumbers
\pagestyle{headings}
\mainmatter

\title{Commands 4 Autonomous Vehicles (C4AV) Workshop Summary}
\titlerunning{Commands 4 Autonomous Vehicles}
\authorrunning{Deruyttere T., Vandenhende S., Grujicic D. et al.}

\author{Thierry Deruyttere$^{1,}$\thanks{Authors contributed equally.} \quad Simon Vandenhende$^{2,}$\samethanks[1] \quad Dusan Grujicic$^{2,}$\samethanks[1] \quad Yu Liu$^2$ \quad Luc Van Gool$^2$ \quad Matthew Blaschko$^2$ \quad Tinne Tuytelaars$^2$ \quad \\ Marie-Francine Moens$^1$}
\institute{KU Leuven\\
$^1$Department of Computer Science (CS)  \\\quad $^2$Department of Electrical Engineering (ESAT) \\
{$^1$\tt\small \{thierry.deruyttere, sien.moens\}@cs.kuleuven.be}\\
{$^2$\tt\small \{firstname.lastname\}@esat.kuleuven.be}}

\maketitle

\begin{abstract}
The task of visual grounding requires locating the most relevant region or object in an image, given a natural language query. So far, progress on this task was mostly measured on curated datasets, which are not always representative of human spoken language. In this work, we deviate from recent, popular task settings and consider the problem under an autonomous vehicle scenario. In particular, we consider a situation where passengers can give free-form natural language commands to a vehicle which can be associated with an object in the street scene. To stimulate research on this topic, we have organized the \emph{Commands for Autonomous Vehicles} (C4AV) challenge based on the recent \emph{Talk2Car} dataset. This paper presents the results of the challenge. First, we compare the used benchmark against existing datasets for visual grounding. Second, we identify the aspects that render top-performing models successful, and relate them to existing state-of-the-art models for visual grounding, in addition to detecting potential failure cases by evaluating on carefully selected subsets. Finally, we discuss several possibilities for future work. 
\end{abstract}

%%%% INTRODUCTION %%%%
\section{Introduction}
The joint understanding of language and vision poses a fundamental challenge for the development of intelligent machines. To address this problem, researchers have studied various related topics such as visual question answering~\cite{antol2015vqa,anderson2018bottom,Johnson2017,Suarez2018,Hudson2018}, image-captioning~\cite{vinyals2015show,xu2015show}, visual grounding~\cite{AAAI20Deruyttere,yu2018mattnet,rohrbach2016grounding}, etc. These advancements can provide a stepping stone towards new products and services. For example, a natural language interface between factory operators and control systems could streamline production processes, resulting in safer and more efficient working environments. In a different vein, providing passengers with the possibility to communicate with their autonomous car could eliminate the unsettling feeling of giving up all control. The possible applications are countless. Understandably, this calls for efficient computational models that can address these tasks in a realistic environment.

In this paper, we focus on the task of visual grounding. Under this setup, the model is tasked with locating the most relevant object or region in an image based on a given natural language query. Several approaches~\cite{AAAI20Deruyttere,yu2018mattnet,rohrbach2016grounding,wang2018learning,hu2016natural,karpathy2014deep} tackled the problem using a two-stage pipeline, where region proposals are generated first by an off-the-shelf object detector~\cite{Faster-RCNN,redmon2018yolov3}, and then matched with an embedding of the sentence. Others~\cite{hudson2018compositional,STACK} proposed an end-to-end strategy where the object location is predicted directly from the input image.

\begin{figure*}
\begin{center}
\scalebox{.9}{
\subfloat[][\textbf{ReferIt}\\ right rocks \\ rocks along the right side \\ stone right side of stairs]{\includegraphics[width=.32\linewidth]{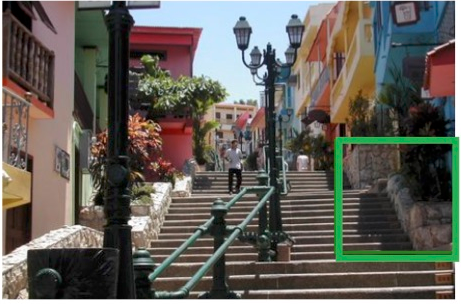}}
\hfill
\subfloat[][\textbf{RefCOCO}\\ woman in white shirt \\ woman on right \\ right woman]{\includegraphics[width=.32\linewidth]{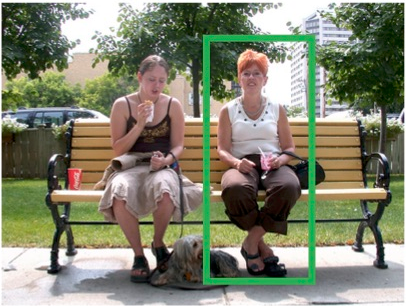}}
\hfill
\subfloat[][\textbf{RefCOCO+}\\ guy in yellow dribbling ball\\ yellow shirt black shorts \\ yellow shirt in focus]{\includegraphics[width=.32\linewidth]{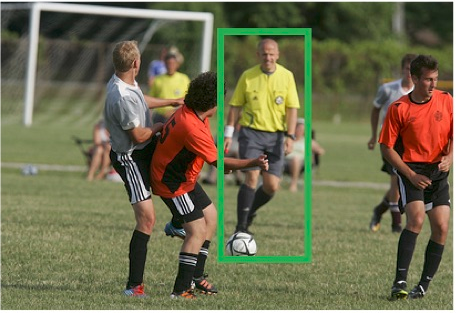}}
}
\end{center}
\caption{Examples from popular benchmarks for visual grounding~\cite{ReferIt}.}
\label{fig:exampleOtherDatasets}
\end{figure*}

In order to quantify progress, several benchmarks were introduced~\cite{ReferIt,yu2016modeling,mao2016generation,hu2018explainable}. From visualizing examples found in existing datasets in Figure~\ref{fig:exampleOtherDatasets}, we draw the following conclusions. First, we observe that the language queries are rather artificial, and do not accurately reflect the type of language used by human speakers during their daily routines. For example, in practice, object references are often implicitly defined, complex and long sentences can contain co-referent phrases, etc. Second, existing benchmarks are mostly built on web-based image datasets~\cite{MSCOCO,hodosh2013framing,young2014image}, where the object of interest is often clearly visible due to its discriminative visual features. From these observations, we conclude that existing benchmarks for visual grounding are not well suited to develop models that need to operate in the wild. 

To address these shortcomings, we hosted the \emph{Commands For Autonomous Vehicles Challenge} (C4AV) at the European Conference on Computer Vision (ECCV) 20'. The challenge setting considers a visual grounding task under a self-driving car scenario. More specifically, a passenger gives a natural language command to express an action that needs to be taken by the autonomous vehicle (AV). The model is tasked with visually grounding the object that the command is referring to. The recently proposed \emph{Talk2Car} dataset \cite{deruyttere2019talk2car} is used to run the challenge. Some examples are displayed in Figure~\ref{fig:exampleCommand}. An extensive description of the challenge can be found in Section~\ref{sect:challenge_description}. 

In contrast to existing benchmarks, several additional challenges are encountered on the \emph{Talk2Car} dataset. First, the referred object can be ambiguous (e.g. there are multiple pedestrians in the scene), but can be disambiguated by understanding modifier expressions in language (e.g. the pedestrian wearing the blue shirt). In some cases, the modifier expressions also indicate spatial information. Second, detecting the correct object is challenging both in the language utterance and the urban scene, for example, when dealing with long and complex sentences, and with small objects in the visual scene, respectively. Finally, the model size and the execution time also play an important role under the proposed task setting. 

The contributions of our work are as follows:
\begin{itemize}
    \item We propose the first challenge for grounding commands for self-driving cars in free natural language into the visual context of an urban environment.
    \item We scrutinize the results obtained by top performing teams. In particular, we compare them against several well-known state-of-the-art models, and further evaluate them on carefully selected subsets that address different key aspects of solving the task at hand, in order to identify potential failure cases. 
    \item Finally, we identify several possibilities for future work under the proposed task setting.
\end{itemize}

%%%% RELATED WORK %%%%
\section{Related Work}
This section provides an overview of recent work on visual grounding. First, several methods are discussed, including both region proposal and non-region proposal based strategies. Next, we review existing benchmarks for visual grounding. 

%%%% METHODS %%%%%
\subsection{Methods}
Existing solutions for visual grounding can be subdivided into two main groups of works. In \emph{Region Proposal Based} methods, object proposals are first generated for the image using an off-the-shelf object detector (typically the region proposal network - RPN). Different works have considered how to correctly match the extracted regions with the language query. In \emph{Non-Region Proposal Based} methods, a model reasons over the full image directly, instead of first extracting object proposals. We discuss representative works for both groups next. 

\subsubsection{Region Proposal Network (RPN) Based Methods}
\citet{hu2016natural} train a model to maximize the likelihood of the referring expression for region proposals that match the object of interest. The global context, spatial configuration and local image features are all taken into account. \citet{rohrbach2016grounding} tackle the problem by learning to attend to regions in the image from which the referring expression can be reconstructed. To this end, they serve a visual representation of the attended regions as input to a text-generating RNN. \citet{wang2018learning} learn a joint embedding for image regions and expressions by enforcing proximity between corresponding pairs through a maximum-margin ranking loss. More recently, modular approaches have seen an increase in popularity.  For example, MAttNet~\cite{yu2018mattnet} decomposes the referring expression into three distinct components, i.e. subject appearance, location and spatial relationships. The different components are subsequently matched with the visual representations, and combined to get a score for each region in the image. Similarly, MSRR~\cite{AAAI20Deruyttere} uses separate modules that focus on text, image and spatial location, and ranking of the image regions respectively. Additionally, the predictions of each module are improved in a recursive manner.

\subsubsection{Non-Region Proposal Based Methods}
\citet{hu2018explainable} apply a modular approach directly to the input image. First, they develop a set of modules that each execute a specific task, and return an attention map over image regions. Next, the expression is decomposed into sub-parts using attention. The extracted parts are considered as sub-problems that can be tackled by the smaller modules learned during the first step. Finally, the answers of the different sub-modules are integrated through an attention mechanism. In contrast to region proposal based methods, the image is subdivided into a 2-dimensional grid. The model predicts the grid cell containing the center of the referred object together with the bounding box offset. 

Another approach that does not rely on the use of region proposals is the work of \citet{hudson2018compositional}. Although this method was originally developed for visual question answering, \cite{deruyttere2019talk2car} adapted it to tackle the visual grounding task. The model uses a recurrent \emph{MAC cell} to match the natural language command with a global representation of the image. First, the MAC cell decomposes the textual input into a series of reasoning steps. Additionally, the MAC cell uses the decomposed textual input to guide the model to focus on certain parts in the image. Information is passed to the next cell between each of the reasoning steps, allowing the model to represent arbitrarily complex reasoning graphs in a soft sequential manner. 

%%%% Datasets %%%%
\subsection{Datasets}
\begin{figure*}[!ht]
\begin{center}
 \centering
\subfloat[You can park up ahead behind \textbf{the silver car}, next to that lamppost with the orange sign on it]{\includegraphics[width=.32\linewidth]{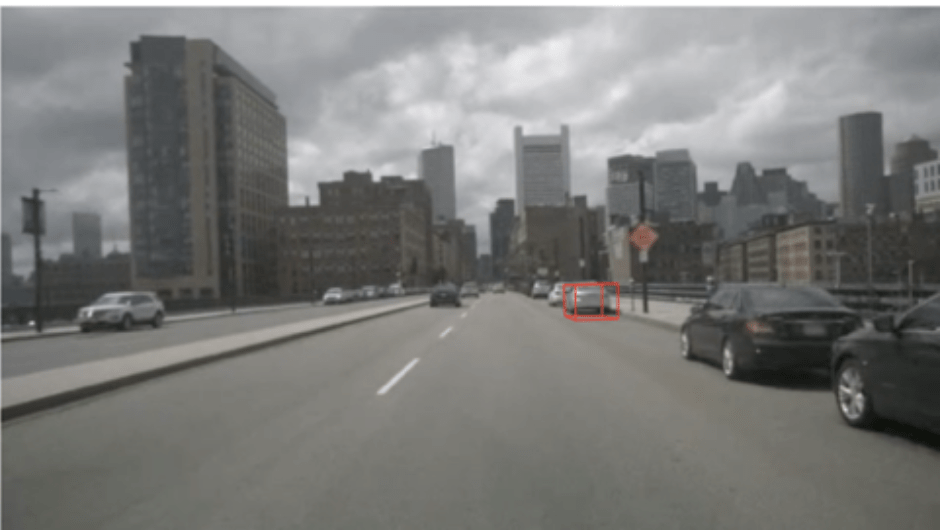}}
\hfill
\subfloat[\textbf{My friend} is getting out of the car. That means we arrived at our destination! Stop and let me out too!]{\includegraphics[width=.32\linewidth]{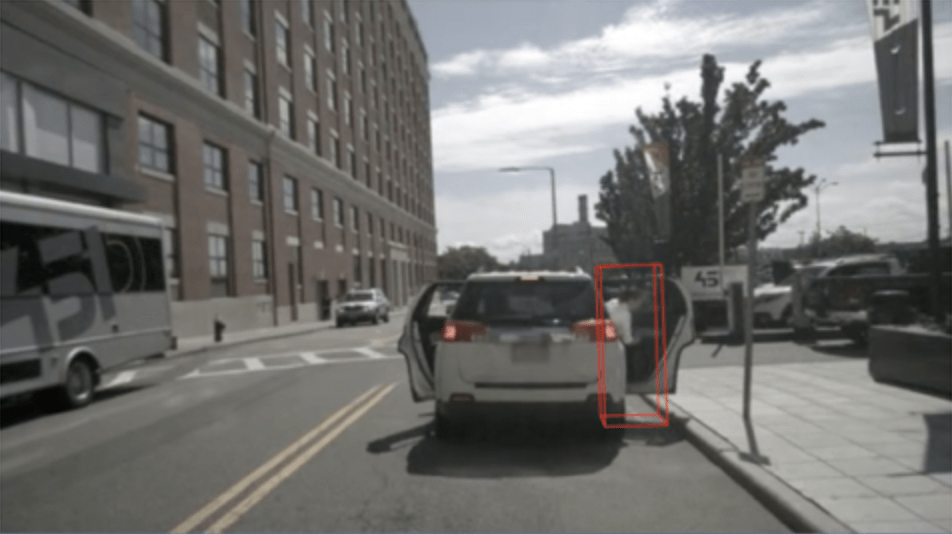}}
    \hfill
\subfloat[Yeah that would be \textbf{my son} on the stairs next to the bus. Pick him up please]{\includegraphics[width=.32\linewidth]{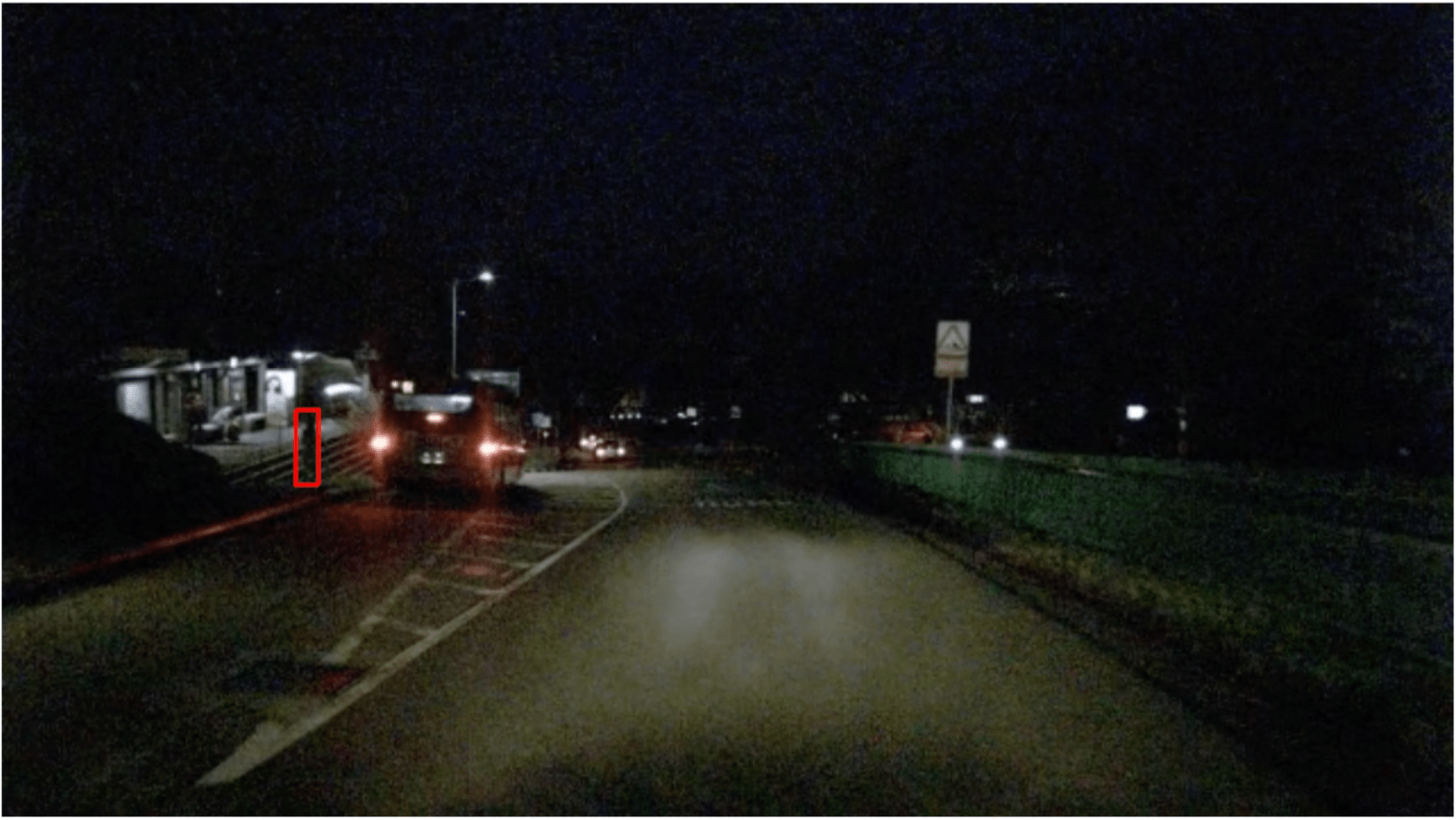}}
\hfill 
\subfloat[After \textbf{that man in the blue top} has passed, turn left]{\includegraphics[width=.32\linewidth]{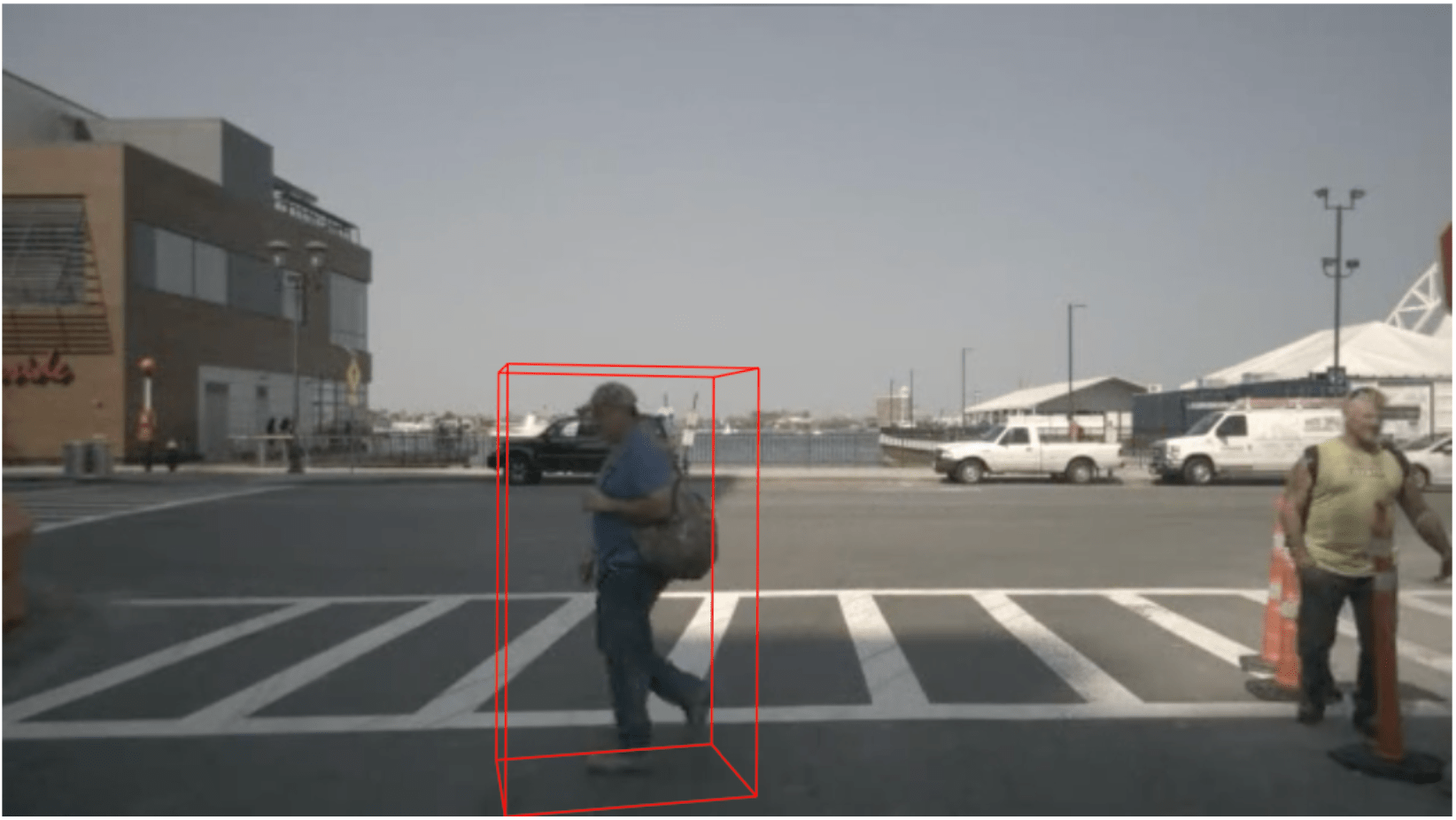}}
    \hfill
\subfloat[There's \textbf{my mum}, on the right! The one walking closest to us. Park near \textbf{her}, she might want a lift]{\includegraphics[width=.32\linewidth]{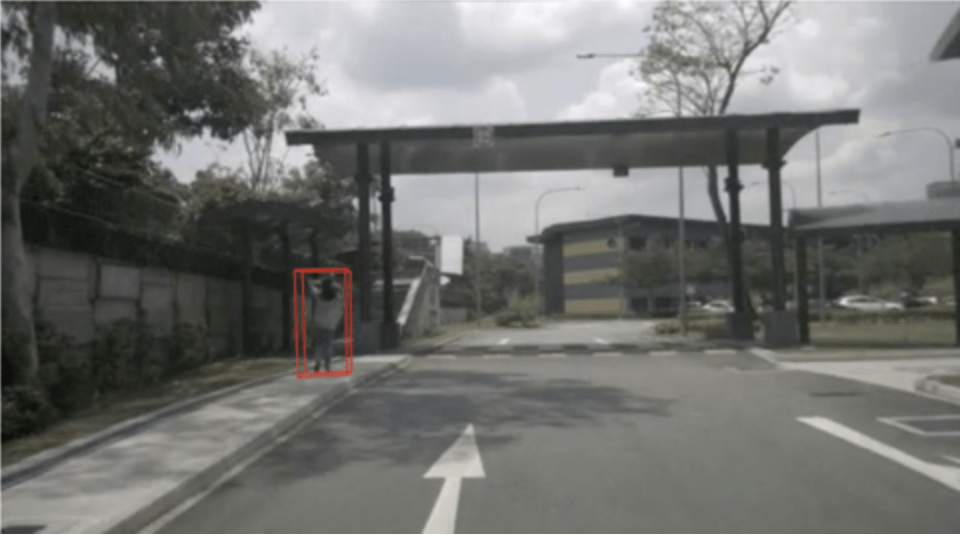}}
\hfill
\subfloat[Turn around and park in front of \textbf{that vehicle in the shade}]{\includegraphics[width=.32\linewidth]{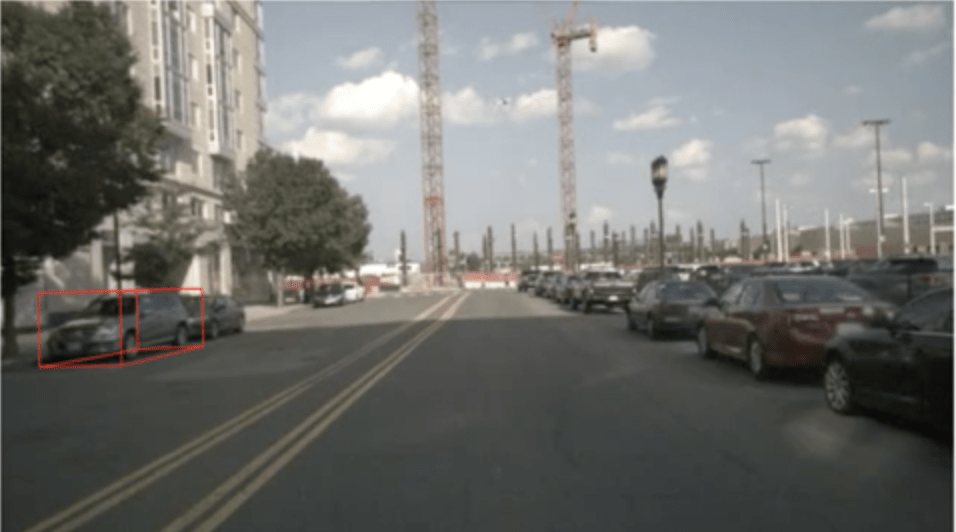}}
\end{center}
   \caption{Some examples from the Talk2Car dataset \cite{deruyttere2019talk2car}. Each command describes an action that the car has to execute relevant to a referred object found in the scene (here indicated by the red 3D-bounding box). The referred object is indicated in bold in each command. Best seen in color.}
  \label{fig:exampleCommand}
\end{figure*}

A number of datasets have been proposed to benchmark progress on the visual grounding task. These include both real-world~\cite{ReferIt,yu2016modeling,mao2016generation} and synthetic~\cite{STACK} datasets. An overview is provided in Table~\ref{tab:comparison_datasets}. Some of the most commonly used datasets are \emph{ReferIt}~\cite{ReferIt}, \emph{RefCOCO}~\cite{yu2016modeling}, \emph{RefCOCO+}~\cite{yu2016modeling} and \emph{RefCOCOg}~\cite{mao2016generation}. These datasets were constructed by adding textual annotations on top of the well-known MS COCO dataset~\cite{MSCOCO}. Examples of image-sentence pairs sampled from these datasets can be seen in Figure~\ref{fig:exampleOtherDatasets}. Notice that the language utterances are rather artificial, i.e. the queries do not accurately present the language used by human speakers. On the other hand, the examples in Figure~\ref{fig:exampleCommand} feature expressions that are more representative of everyday language, e.g. object references are often less explicit, and part of longer and more complex sentences. Additionally, the images found in the aforementioned datasets were collected from the web, and as a consequence, the objects of interest are often quite easy to spot. A different situation arises when considering indoor or urban scene environments. 

In contrast to prior works, \citet{vasudevan2018object} and \citet{deruyttere2019talk2car} considered the visual grounding task in a city environment. The main difference between the two works is the use of object descriptions in the former, versus the use of command-like expressions with more implicit object references in the latter. In this work, we use the \emph{Talk2Car}~\cite{deruyttere2019talk2car} dataset. 

\begin{table*}[ht!]
\small
\begin{center}
\scalebox{0.9}{
\begin{tabular}{|l|c|c|c|c|c|c|c|c|}
\hline
Dataset & Images & Objects & Expressions & Avg expr length  & Video & Lidar & Radar \\
\hline
\texttt{ReferIt} \cite{ReferIt} &  19,894 &  96,654  & 130,525 & 3.46 &$\times$&$\times$&$\times$ \\
\texttt{RefCOCO} \cite{yu2016modeling} &  26,711 & 50,000 & 142,209 & 3.61 &$\times$&$\times$&$\times$ \\
\texttt{RefCOCO+} \cite{yu2016modeling} & 19,992 & 49,856 & 141,564 & 3.53 &$\times$&$\times$&$\times$ \\
\texttt{RefCOCOg} \cite{mao2016generation}& 26,711 & 54,822 & 85,474  & 8.43 &$\times$&$\times$&$\times$ \\
\texttt{CLEVR-Ref} \cite{STACK} & 99,992 & 492,727 & 998,743 & 14.50 &$\times$&$\times$&$\times$  \\ 
\texttt{Cityscapes-Ref} \cite{vasudevan2018object} & 4,818 &  29,901  & 30,000  & 15.59 &\checkmark& $\times$ & $\times$ \\
\texttt{Talk2Car} \cite{deruyttere2019talk2car} & 9,217 & 10,519 & 11,959 & 11.01 &\checkmark &\checkmark &\checkmark  \\
\hline
\end{tabular}
}
\end{center}
\caption{An overview of public datasets for visual grounding \cite{deruyttere2019talk2car}.}
\label{tab:comparison_datasets}
\end{table*}

%%%% CHALLENGE %%%%
\section{Commands for Autonomous Vehicles Challenge}
\label{sect:challenge_description}

This section describes the 'Commands for Autonomous Vehicles' (C4AV) challenge that was hosted as part of the C4AV workshop at ECCV 20'. First, we introduce the used benchmark. Second, we define three baseline models that were provided at the start of the challenge. Finally, we give an overview of the top performing models at the end of the competition. 

\subsection{Dataset}
\label{sect:Talk2Car}
The C4AV challenge is based on the \emph{Talk2Car} dataset~\cite{deruyttere2019talk2car}. The dataset is built on top of the nuScenes~\cite{caesar2020nuscenes} dataset which contains 3D object boxes, videos, lidar and radar data obtained by driving a car through Boston and Singapore. Furthermore, the nuScenes dataset covers various weather conditions, different lighting conditions (day and night), and driving directions (left and right). The \emph{Talk2Car} dataset is constructed by adding textual annotations on top of the images sampled from the nuScenes training dataset. More specifically, the text queries consist of commands provided by a passenger to the autonomous vehicle. Each command can be associated with an object visible in the scene. Figure~\ref{fig:exampleCommand} shows text-image pairs from the \emph{Talk2Car} dataset. The challenge required to predict the 2D bounding box coordinates around the object of interest. Additionally, for every image, we provided pre-computed region proposals extracted with a CenterNet~\cite{zhou2019objects} model. 

The dataset contains $11 959$ text-image pairs in total. The train, val and test set contain $70 \%$, $10 \%$ and $20 \%$ of the samples, respectively. Additionally, the test set can be subdivided into four sub-sets of increasing difficulty. One subset was created to study the specific case of objects that are far away from the vehicle. Two sub-sets contain varying command lengths. The fourth subset tests how well the model disambiguates between objects of the same class as the target. 

The C4AV challenge was hosted on \emph{AICrowd} prior to ECCV 20'. Every team was allowed a maximum of three submissions per day. The submissions were evaluated on the held-out test set from the \emph{Talk2Car} dataset. The evaluation criterion is described in Section~\ref{sect:evaluation}. Top performing teams were invited to submit a paper to the workshop after undergoing a code verification phase. 

\subsection{Baselines}
\label{sect:baselines}

The leader board featured three baseline models at the start of the challenge. We describe each of them below. For brevity, we adopt the following notations. Each model takes as input an image $I$ and a natural language expression under the form of a command $C$. The objective is to localize the referred object $o$. 

\subsubsection{Bi-directional retrieval}
\label{subsubsec: bidirectional}
Similar to the work from~\citet{karpathy2014deep}, a Bi-directional retrieval model is considered. First, a pre-trained object detector~\cite{zhou2019objects} is used to generate region proposals from the image $I$. Second, a pre-trained ResNet-18 model is used to obtain a local feature representation for the extracted regions. Similarly, the command $C$ is encoded by a bi-directional GRU. Finally, we match the command $C$ with the correct region proposal. To this end, we maximize the inner product between the local image features of the correct region and the text encoding, while we minimize the inner product for the other regions. To help participants get started in the challenge, a PyTorch~\cite{paszke2019pytorch} implementation of the Bi-directional retrieval model was made publicly available~\cite{vandenhende2020baseline}.

\subsubsection{MAC}
As a second baseline, we consider the MAC model~\cite{Hudson2018}, adapted for the visual grounding task by~\citet{deruyttere2019talk2car}. MAC implements a multi-step reasoning approach, and unlike the Bi-directional retrieval model, does not rely on pre-computed region proposals. First, the image is encoded by a pre-trained ResNet model~\cite{ResNet}, and the command tokens are encoded by a bidirectional LSTM. Second, the model initiates a multi-step reasoning process by applying attention to the command tokens. In each reasoning step, the model attends to a different word, decomposing the language query into smaller sub-problems. Simultaneously, visual information is extracted from the image through a soft-attention mechanism conditioned on the attended word. The extracted information is stored as a memory vector, and forwarded to the next reasoning step. The final product of the multi-step reasoning process is a 2D bounding box derived from the soft-attention mask applied to the visual features. 

\subsubsection{MSRR}
As a final baseline model, we consider the prior state-of-the-art on the \emph{Talk2Car} dataset. Similar to MAC~\cite{Hudson2018}, the MSRR model~\cite{AAAI20Deruyttere} employs a multi-step reasoning strategy. However, unlike MAC, MSRR obtains its predictions by ranking object region proposals as in Section~\ref{subsubsec: bidirectional}. First, each region is associated with a separate spatial map for which we indicate the spatial location by assigning ones to the regions, while zeros otherwise. Next, MSRR decomposes the expression $C$ into sub-parts through attention and a region dependent reasoning process. The latter is achieved by (i) multiplying the spatial location of each region and its score with the image features, extracted by a ResNet model, and (ii) applying soft-attention conditioned onto the decomposed sub-expression. The result is combined into a region specific memory vector which can be used to score the region based on the alignment of the newly created memory vector and the command $c$. Finally, the highest scoring region is returned. 

\subsection{State-of-the-art models}
\label{sect:sota_models}
Teams that outperformed all three baselines at the end of the challenge were invited to submit a paper to the workshop detailing their solution. A selection of top performing models is summarized below.

\subsubsection{Stacked VLBert}
\citet{stacked_vlbert} propose a visual-linguistic BERT model named \emph{Stacked VLBert}. The approach relies on region proposals, similarly to the MSRR \cite{AAAI20Deruyttere} and the Bi-directional retrieval model \cite{vandenhende2020baseline}. Furthermore, the authors propose a weight stacking method to efficiently train a larger model from a shallow VLBert model. The weight stacking procedure is performed by first training a smaller VLBert variant and then copying its trained weights in a repeated manner into a larger model. They show that by doing this, they can achieve a higher score compared to a larger model initialised with random weights. The Stacked VLBert model is conceptually close to the Bi-directional retrieval model as they first encode the image and object regions with a pre-trained image encoder. They then pass the sentence, the encoded image, and the encoded objects to the VLBert model to find the referred object.

\subsubsection{Cross-Modal Representations from Transformers (CMRT)}
% \citet{CMRT} also advocate the use of a region proposal based approach. More specifically, the CMRT model refines the basic Bi-directional retrieval model. The image and command are encoded using an off-the-shelf image encoder and transformer model respectively. Afterwards, the feature representations are aggregated and refined using another transformer model. Unlike the basic Bi-directional retrieval model, the encoding of the complete image is considered. This allows to capture long-range dependencies across the image. The latter is important since the commands on the \emph{Talk2Car} dataset also include the surrounding in the description of the objects. After aggregating the feature representations, an RoI alignment operation is used to select local image features from the regions of interest. The cropped features are fed to a final weight sharing network which is optimized using the same objective function as the Bi-directional retrieval model.

\citet{CMRT} also advocate the use of a region proposal based approach. The commands are fed to the input of a transformer encoder, while the image features are used as the query for the transformer decoder. The image features are refined based on the extracted linguistic features obtained from the encoder, which are used as the key and value input to the multi-head attention layers in the decoder. Unlike the common approaches that leverage the transformer encoder alone to extract visual-linguistic features, CMRT uses the transformer decoder to aggregate features from the two modalities. Additionally, as opposed to extracting local features from the region crops, the features of the whole image are used as the decoder input. This allows to capture long-range dependencies in the image, which is important, since the \emph{Talk2Car} dataset commands also include the surroundings in the description of the objects. After aggregating the feature representations and extracting the feature map at the decoder output, an RoI alignment operation is used to select local image features of interest. The cropped features are fed to a final weight sharing network which is optimized using the same objective function as the Bi-directional retrieval model.

% \subsubsection{Cosine meets Softmax: A tough-to-beat baseline
% for visual grounding (CMSVG)}
% \citet{CMSVG} showed that the Bi-directional retrieval approach can outperform more complicated competitors such as MSRR~\cite{AAAI20Deruyttere} and MAC~\cite{Hudson2018} when applying sufficient finetuning. In particular, they performed extensive ablation studies to analyse the influence of the used number of region proposals, the used image encoder, and the used text encoder. The obtained results suggest that using a better image or text encoder provides a more viable alternative to boost the performance on the visual grounding task, compared to using a conceptually more complex approach based on multi-step reasoning.

\subsubsection{Cosine meets Softmax: A tough-to-beat baseline
for visual grounding (CMSVG)}
\citet{CMSVG} showed that the Bi-directional retrieval approach can outperform more sophisticated approaches such as MSRR~\cite{AAAI20Deruyttere} and MAC~\cite{Hudson2018} %when applying sufficient finetuning. 
by simply using state-of-the-art object and sentence encoders. They also performed extensive ablation studies to analyse the influence of the used number of region proposals, the used image encoder, and the used text encoder.
%In particular, they performed extensive ablation studies to analyse the influence of the used number of region proposals, the used image encoder, and the used text encoder. The obtained results suggest that, for tackling the visual grounding task, using a simple setup similar to that of the Bi-directional retrieval model, but with a better image or text encoder provides a viable alternative to the conceptually more complex approaches as MAC \cite{hudson2018compositional} and MSRR \cite{AAAI20Deruyttere}.%based on multi-step reasoning or attention mechanisms.

\subsubsection{Attention Enhanced Single Stage Multi-Modal Reasoner (ASSMR)}
~\citet{ASSMR} also encode the local information of pre-computed region proposals first. Additionally, the position and scale of the region proposals are encoded and supplied as extra information for every region. Note that these properties are often found in the modifier expressions of the commands, and are thus potentially informative of the object location. Next, the local image features are combined with an encoding of the command to compute the weight for every region, emphasizing the ones most relevant for the given command. Afterwards, the weighted region features are augmented with a global image representation, and aggregated with the hidden states of a GRU that was used to encode the command. An attention mechanism is applied to the obtained multi-modal feature representation and the local object features to compute a final score for each region. 

\subsubsection{AttnGrounder: Talking to Cars with Attention}~\citet{AttnGrounder} proposes a one-stage approach to the visual grounding task. A Darknet-53 \cite{redmon2018yolov3} backbone is utilized for extracting image features at multiple spatial resolution, while a bidirectional LSTM is used to generate text features. A visual-text attention module that  relates  every  word  in  the  given  query  with different image regions is used to construct a unique text representation for each region. Additionally, the prediction of a segmentation mask within the bounding box of the referred object is introduced as an auxiliary task, improving the localization performance. The predicted mask serves as an attention map used to weigh the visual features, which are subsequently concatenated with the spatially attended text features and the original visual features along the channel dimension, and fused together by using 1x1 convolutions. Finally, similarly to YOLOv3 \cite{redmon2018yolov3}, the fused features are used to predict the offset from anchor boxes and categorical labels that indicate whether the predicted bounding box corresponds to the ground truth.

\section{Evaluation}
\label{sect:evaluation}
Every model needs to output a 2D bounding box that indicates the location of the referred object. The challenge submissions were ranked using the evaluation measure from~\citet{deruyttere2019talk2car}. In particular, we employed the $IoU_{.5}$ metric by thresholding the Intersection over Union (IoU) between the predicted and ground-truth bounding boxes at $0.5$. The IoU is defined as follows:
\begin{equation}
IoU = \frac{\text{Area of Overlap of the two boxes}}{\text{Area of Union of the two boxes}}.
\end{equation}

While the challenge focused on the quality of the predictions, other properties such as model size and inference speed are arguably important as well in our task setting. To draw attention to these problems, for every model, we also report the number of parameters and the inference speed on a Nvidia RTX Titan 2080. 

%%%%%% EXPERIMENTS
\section{Experiments}

This section analyses the results obtained by the models described in Section~\ref{sect:baselines} and~\ref{sect:sota_models}. In particular, Section~\ref{sect:sota_comparison} draws a comparison between the state-of-the-art on the \emph{Talk2Car} dataset. Section~\ref{sect:subsets} evaluates the models on carefully selected subsets to better understand any existing failure cases. Finally, Section~\ref{sect:qualitative_comparison} analyses the commonalities and differences between the used models to isolate the elements that render top-performing models successful.

\subsection{State-Of-The-Art Comparison}
\label{sect:sota_comparison}

Table~\ref{tab:t2c_test_set_results} compares the models from Section~\ref{sect:baselines} and~\ref{sect:sota_models} on the \emph{Talk2Car} test set. We compare the models in terms of $IoU_{.5}$, the number of parameters in millions (M) and inference speed in milliseconds (ms). The state-of-the-art prior to the challenge was the MSRR model~\cite{AAAI20Deruyttere}. Notably, the top-performing models from the challenge% leader board
~\cite{stacked_vlbert,CMRT,CMSVG,ASSMR,AttnGrounder} show significant gains over the prior state-of-the-art~\cite{AAAI20Deruyttere} in terms of performance ($IoU_{.5}$). In particular, the Stacked VLBert model establishes a new state-of-the-art, and outperforms prior work by $10.9\%$ $IoU_{.5}$. Furthermore, %looking closer 
we find that some of the top-performing models~\cite{stacked_vlbert,CMSVG} drew a lot of inspiration from the Bi-directional retrieval model~\cite{vandenhende2020baseline}. We conclude that when using strong visual and textual feature representations, this simple model can outperform more complex schemes like MAC~\cite{Hudson2018} and MSRR~\cite{AAAI20Deruyttere}. 

\noindent\textbf{Extra Test} To verify the validity of the predictions on the leader board, we evaluated the models on an additional hidden test set of 100 commands after the challenge (see $IoU_{.5} \dagger$ in Table~\ref{tab:t2c_test_set_results}). None of the submissions experiences a significant performance drop compared to the results on the official test set (see $IoU_{.5}$ vs  $IoU_{.5} \dagger$). This confirms the validity of the models. The additional test set annotations will be released after the workshop. 

\noindent\textbf{Resource Analysis} Although the challenge focuses on the quality of the predictions, the used amount of computational resources needs careful consideration too. We performed a detailed resource analysis in Table~\ref{tab:t2c_test_set_results}. Except for ASSMR~\cite{ASSMR}, models that improve over MSRR~\cite{AAAI20Deruyttere} do this at the cost of increasing the model size (parameters). However, we do see improvement in terms of inference speed. The advantage of using a simple bi-directional retrieval approach over a more complex multi-step reasoning process is clearly visible here. 

\begin{table}[]
    \centering
    \begin{tabular}{l|c|c|c|c}
        Model & $IoU_{.5}$ & $IoU_{.5}$ $\dagger$ & Params (M) & Inference Speed (ms)  \\
        \hline
        Stacked VLBert \cite{stacked_vlbert} & \textbf{0.710} & \textbf{0.762} & 683.80 & 320.79 \\
        CMRT \cite{CMRT} & $0.691$ & 0.713 & 194.97 & 215.50 \\
        CMSVG \cite{CMSVG} & $0.686$ & 0.733 & 366.50 & 164.44 \\
        ASSMR \cite{ASSMR} & $0.660$ & 0.723 &\textbf{48.91} & 126.23 \\
        AttnGrounder \cite{AttnGrounder} & $0.633$ & $0.613$ & 75.84 & \textbf{25.50} \\
        \hline
        MSRR \cite{AAAI20Deruyttere} & $0.601$ & 0.634&  62.25 & 270.50  \\
        MAC \cite{hudson2018compositional} & $0.505$ & 0.525 & 41.59 & 51.23 \\
        Bi-Directional retr. \cite{vandenhende2020baseline} & $0.441$ & 0.327 & 15.80 & 100.24 \\
    \end{tabular}
    \caption{Results on the \emph{Talk2Car} test set. The inference speed was measured on a single Nvidia RTX Titan. $\dagger$ Results on an extra smaller test set that was hidden from the leaderboard.}
    %to verify correctness of the results.}
    \label{tab:t2c_test_set_results}
\end{table}

\subsection{Talk2Car Subsets}
\label{sect:subsets}

\begin{figure*}[t]
\begin{center}
\includegraphics[width=1\linewidth]{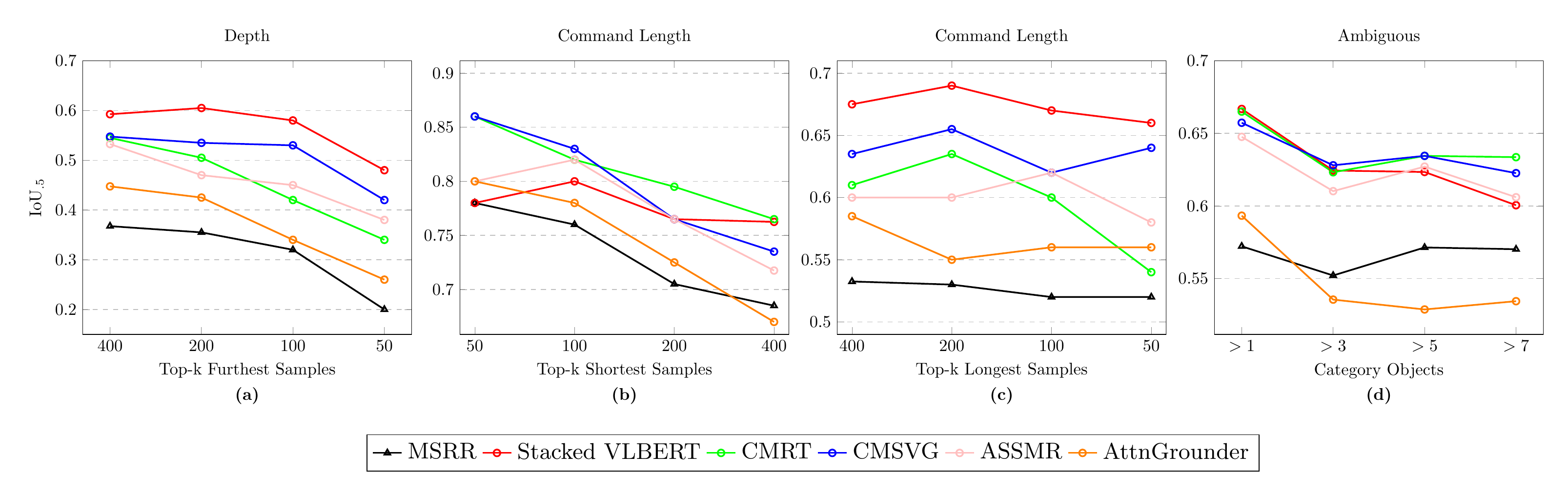}
%\includestandalone[width=1\linewidth]{plots}
\caption{
Results of the state-of-the-art models (Section~\ref{sect:sota_models} on four sub-test sets from the \emph{Talk2Car} dataset. Each plot shows the easy examples on the left, while the difficulty increases as we move to the right along the axis. In plot (a) - (c), we increase the difficulty by choosing the top-$k$ samples under the selected criterion (e.g. depth, expression length). In plot (d), we increase the difficulty by choosing samples for which multiple same-category objects are present in the scene.} 
\label{fig:subset_tests}
\end{center}
\end{figure*}

Section~\ref{sect:Talk2Car} described the construction of four carefully selected smaller test sets on the \emph{Talk2Car} benchmark. In this section, we evaluate the models under various challenging conditions using the different subsets. Interestingly, the level of difficulty can be tuned as well. As a concrete example, the first subset is constructed by selecting the top-$k$ examples from the dataset for which the object of interest is furthest away. By reducing the value of $k$, we effectively test on objects that are harder to spot due to their increased distance from the vehicle. For each subset, we consider four levels of difficulty. Figure~\ref{fig:subset_tests} shows the results on the subsets. Next, we consider each of the four subsets in detail. 

\noindent\textbf{Far Away Objects} Figure~\ref{fig:subset_tests}(a) shows the results when focusing on far away objects. All models follow a similar trend, i.e. the performance drops significantly for objects that are further away. This observation follows from the use of the CenterNet~\cite{zhou2019objects} region proposals by all models. Notice that far away objects often tend to be small in size. The detection of small objects has been studied by several works, and requires specific dedicated solutions~\cite{chen2016r,li2017perceptual}.

\noindent\textbf{Short commands}
Figure~\ref{fig:subset_tests}(b) displays the results when increasing the length of the commands. All models tend to score higher on the commands that are shortest in length. Yet, the performance drops significantly when increasing the length of the commands, i.e. going from the top-50 to the top-400 shortest commands. This is surprising since the maximum sentence length only sees a small increase, i.e. from 4 (top-50) to 6 (top-400). We believe that it would be useful to study the reason behind this behavior in future work. 

\noindent\textbf{Long commands} Figure~\ref{fig:subset_tests}(c) measures the performance on the image-sentence pairs with the top-$k$ longest commands. Depending on the used model, the performance is more susceptible to commands of increasing length. In particular, CMRT~\cite{CMRT} shows a decline in performance, while the performance of the other models remains more or less constant (less than $1.5\%$ difference). We hypothesize that the transformer model~\cite{CMRT} responsible for aggregating the image and sentence features does not perform well in combination with long sentences. 

\noindent\textbf{Ambiguity} Finally, Figure~\ref{fig:subset_tests}(d) shows the performance when considering scenes with an increasing number of objects of the same category as the referred object. The CMRT~\cite{CMRT} model obtains the highest performance when considering scenes with many ambiguous objects ($> 7$). In this case, the object of interest can only be identified through its spatial relationship with other objects in the scene. In this case, the global image representation used by the CMRT model is beneficial. Similarly, the MSRR model~\cite{AAAI20Deruyttere} handles the ambiguous scenes rather well. This can be attributed to its multi-step reasoning process, taking into account all objects and their spatial relationships. The AttnGrounder~\cite{AttnGrounder}, on the other hand, experiences a significant performance drop on ambiguous cases, potentially due to the fact that it does not utilize region proposals. Appendix~\ref{subsect:MSRR_Reasoning} shows a successful case where successive reasoning steps are beneficial. 

\subsection{Qualitative comparison}
\label{sect:qualitative_comparison}

Finally, Table~\ref{tab:insight_table} gives a qualitative overview of all the methods under consideration. In particular, we consider the following elements: model performance, visual backbone, language model, word attention, image augmentations, language augmentations, region proposal based vs non-region proposal based and whether a global image representation is used or not. Based on this comparison, we present some additional findings below. 

\noindent\textbf{Region Proposal Networks} It is worth noting that all the top-performing models, except for the AttnGrounder~\cite{AttnGrounder}, are based on pre-computed region proposals. We hypothesize that doing so is particularly interesting under the \emph{Talk2Car} setting since the scenes are often cluttered with information. This is corroborated by the fact that the models that utilize the region proposals perform better on samples with a large number of objects from the target object category (which often implies a large number of objects in general), as can be seen on Figure~\ref{fig:subset_tests}. However, there is a downside to the use of region proposals as well. For example, the model can not recover when the region proposal network fails to return a bounding box for the object of interest. In this case, relative regression or query-based approaches can be used~\cite{Chen_2017_ICCV, kovvuri2018pirc}.

\noindent\textbf{Backbones and word attention} Surprisingly, some of the higher ranked methods use the same visual backbone as their lower ranked competitors. On the other hand, higher ranked models can be associated with more recent, better performing language embeddings. We conclude that the effect of using a deeper or better image encoder is smaller compared to using a better linguistic representation. Furthermore, word attention seems to be an additional important contributing factor in state-of-the-art models, %. %Evidently, there is merit in the use of word attention 
as it allows them to focus on key words. %, while ignoring less informative sentence parts.

\noindent\textbf{Augmentations} 
Remarkably, some models do not use augmentations although it has been shown that these are important for image recognition tasks~\cite{wvangansbeke2020learning,sohn2020fixmatch}.
It is also not surprising that the use of augmentations on the language side has not really been explored, as it is rather difficult. We are interested to see if recent frameworks like~\cite{ma2019nlpaug} can prove helpful in this case. 

\noindent\textbf{Global Image Representation} Finally, we consider whether a global image representation is taken into account or not. Note that this is optional in case of region proposal based methods. Apart from the Bi-directional retrieval baseline and CMSVG~\cite{CMSVG}, all models used global context information to make the predictions. This observation confirms that it is beneficial to use global image information. Doing so allows to better capture the spatial relationships between objects. These are likely important to tackle the C4AV challenge. 

\begin{sidewaystable}
\begin{tabular}{l|c|c|c|c|p{2.5cm}|p{2.5cm}|c|c}
    Model & $IoU_{.5}$ & Vision Backb. & Language Backb. & Word Att. &  Vision Augm. & NLP Augm. & RPN & Global Image repr. \\
    \hline
    Stacked VLBert \cite{stacked_vlbert} & \textbf{0.710} & ResNet-101~\cite{ResNet}  & VLBERT~\cite{Su2020VL-BERT:} & Yes & %HZ Flip \newline Affine Transf. & ``Left''$\xrightarrow[]{}$``Right'' or vice versa when horiz. flip 
    Undisclosed & Undisclosed & Yes & Yes \\
    \hline
    CMRT \cite{CMRT} & $0.691$ & ResNet-152~\cite{ResNet} & Transformer~\cite{vaswani2017attention} & Yes & 
    Undisclosed & Undisclosed & Yes &  Yes \\
    \hline
    CMSVG \cite{CMSVG} & $0.686$ & EfficientNet~\cite{tan2019efficientnet} & Sent.-Transf. \cite{reimers-2019-sentence-bert} & Yes & No & No & Yes & No \\
    \hline
    ASSMR \cite{ASSMR} & $0.660$ & ResNet-18~\cite{ResNet} & GRU & Yes & HZ Flip \newline Rotation \newline Color Jitter & No & Yes & Yes \\
    \hline
    AttnGrounder \cite{AttnGrounder} & $0.633$ & Darknet-53 & LSTM & Yes & HZ Flip \newline Random Affine  \newline Color Jitter & No & No & Yes \\
    \hline
    MSRR \cite{AAAI20Deruyttere} & $0.601$ & ResNet-101~\cite{ResNet} & LSTM  & Yes & No & No & Yes & Yes \\
    \hline
    MAC \cite{hudson2018compositional} & $0.505$ & ResNet-101~\cite{ResNet} & LSTM & Yes & No & No & No & Yes \\
    \hline
    Bi-Dir. retr. \cite{vandenhende2020baseline} & $0.441$ & ResNet-18~\cite{ResNet} & LSTM & No & No & No & Yes & No \\
\end{tabular}
 \caption{Qualitative comparison of all methods under consideration. A description of every model can be found in Section~\ref{sect:baselines} and~\ref{sect:sota_models}.}
 \label{tab:insight_table}
\end{sidewaystable}

%%%%% CONCLUSION
\section{Conclusion}

In this paper, we reviewed the results of the \emph{Commands for Autonomous Vehicles} challenge held at ECCV20'. First, we presented an overview of various strategies to tackle the visual grounding task. For each method, we described its key components, and discussed the commonalities and differences with existing works. Second, we presented an extensive experimental evaluation of the considered methods. We briefly discuss some of the limitations and possibilities for future work. 

\noindent\textbf{Dataset} The \emph{Talk2Car} dataset provides a more realistic task setting compared to existing benchmarks~\cite{ReferIt,yu2016modeling,mao2016generation}, yet we identify several possibilities to extend this work. First, the dataset could be further up-scaled in terms of the number of commands, and the variety of the environments. Second, it would be interesting to add annotations containing novel classes~\cite{sadhu2019zeroshot} and groups of objects. 

\noindent\textbf{Extra Modalities} Surprisingly, the use of other sensor modalities, e.g. depth, LIDAR, maps, etc., remains unexplored. Still, this provides an interesting direction for future research. In particular, leveraging additional data sources as extra input~\cite{gupta2014learning} or auxiliary task~\cite{vandenhende2020revisiting} is expected to boost the performance. 

\noindent\textbf{Navigation} The current setup considers the task of visual grounding in isolation. Yet, the agent is also responsible for navigating to the correct destination. Extending the current setup with a navigation task~\cite{anderson2018vision,savva2019habitat,vasudevan2019talk2nav,thomason2020vision,chen2019touchdown} would provide a useful addition.  

\section{Acknowledgements}
This project is sponsored by the MACCHINA project from the KU Leuven with grant number C14/18/065. Additionally, we acknowledge support by the Flemish Government under the Artificial Intelligence
(AI) Flanders programme. Finally, we thank Huawei for sponsoring the workshop and AICrowd for hosting our challenge. 
\clearpage

%\bibliographystyle{splncs}
%\bibliography{egbib}

\newpage
\begin{subappendices}
\renewcommand{\thesection}{\Alph{section}}%
% or try \arabic{section}

\section{}
\subsection{Multi-step Reasoning MSRR}
\label{subsect:MSRR_Reasoning}

This section discusses the influence of having reasoning steps and showcases an example where the MSRR \cite{AAAI20Deruyttere} successfully finds the correct answer for the command by using multiple reasoning steps.

\begin{figure*}[h!]
  \centering
    \includegraphics[width=\textwidth]{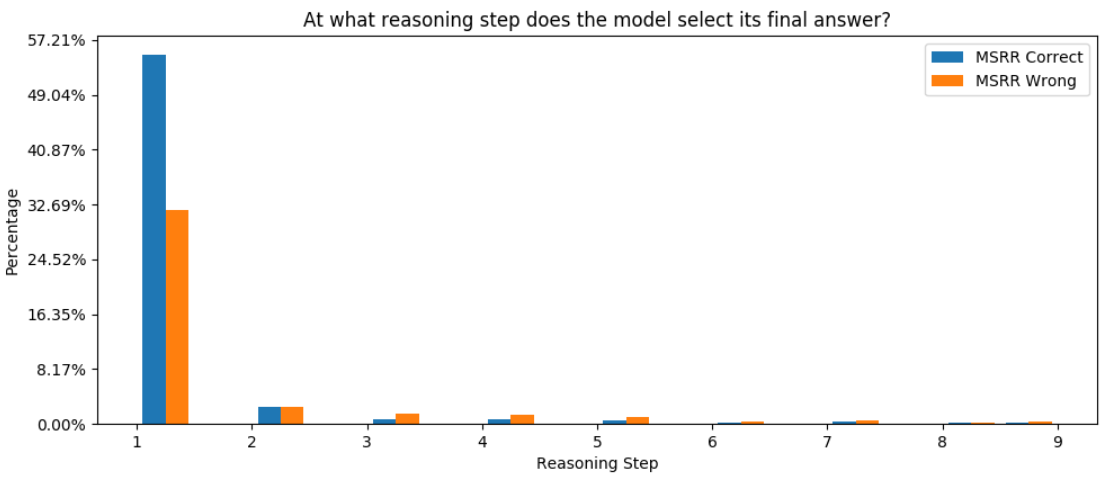}
\caption{This plot shows in which step a 10 reasoning step MSRR makes its final decision. We use MSRR Correct (blue) to indicate when the final answer by the model is also the correct answer while MSRR Wrong (orange) is used when the final answer is the wrong answer.}
    \label{fig:MSRR_change_decision}
\end{figure*}

First, we will look at the influence of reasoning steps.
Assume we have a MSRR model that uses 10 reasoning steps, Figure \ref{fig:MSRR_change_decision} shows in which of these 10 reasoning steps the model makes its final prediction. It is clear that most of the final predictions are made in the very first reasoning step. For instance, if we would only consider the answers in the first step and ignore any change of decision in the following steps, we would achieve $\approx$ 55\% $IoU_{.5}$. Yet, by including more reasoning steps we can further improve this to $\approx$ 60\% $IoU_{.5}$. This shows that having reasoning steps can be beneficial for this kind of task.

\begin{figure*}[h!]
  \centering
    \includegraphics[width=\textwidth]{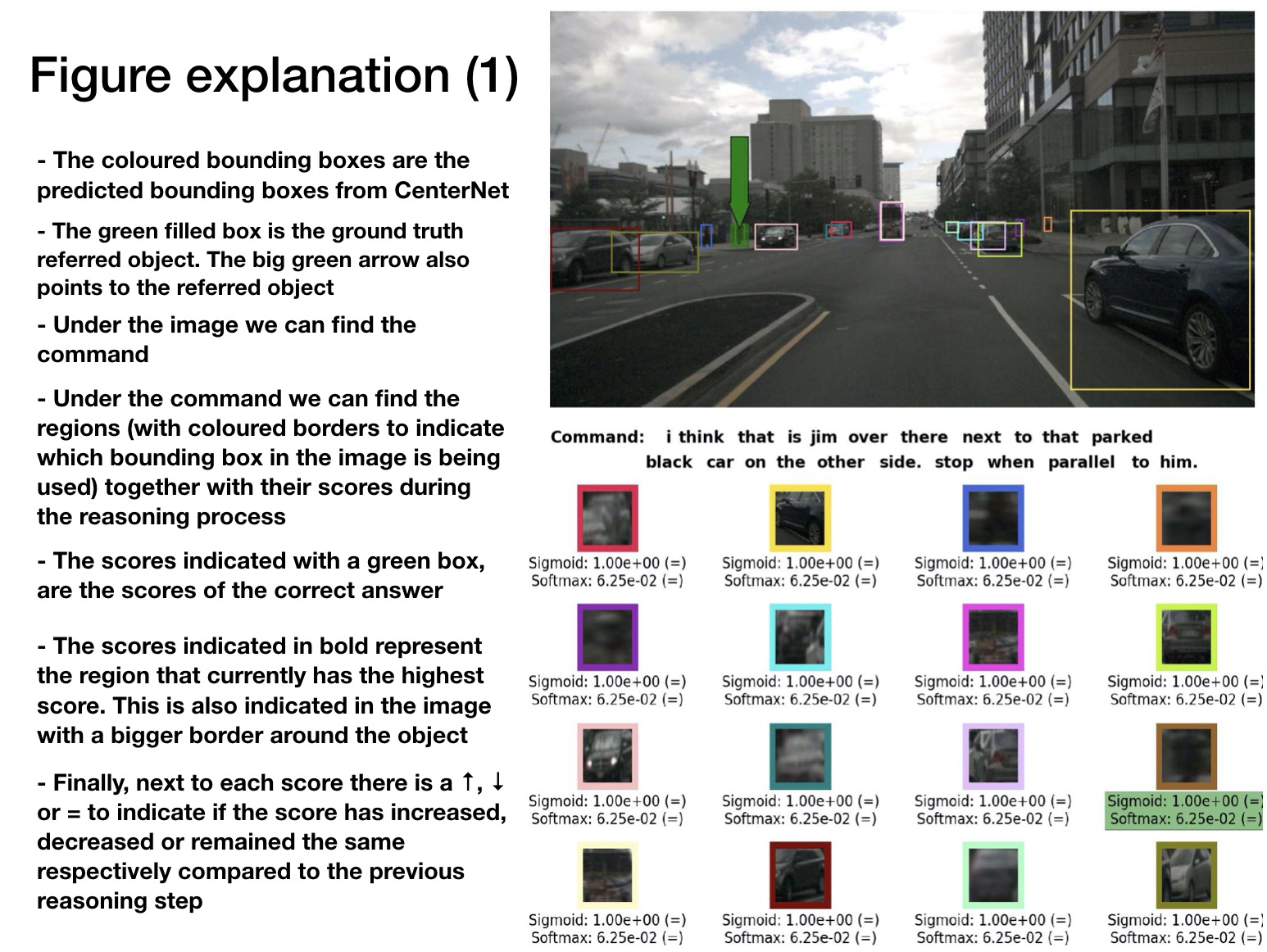}
\caption{Explaining the visualisation of the reasoning process  (Part 1). Figure from \cite{AAAI20Deruyttere}.
}
    \label{fig:vis_expl1}
\end{figure*}

\begin{figure*}[h!]
  \centering
    \includegraphics[width=\textwidth]{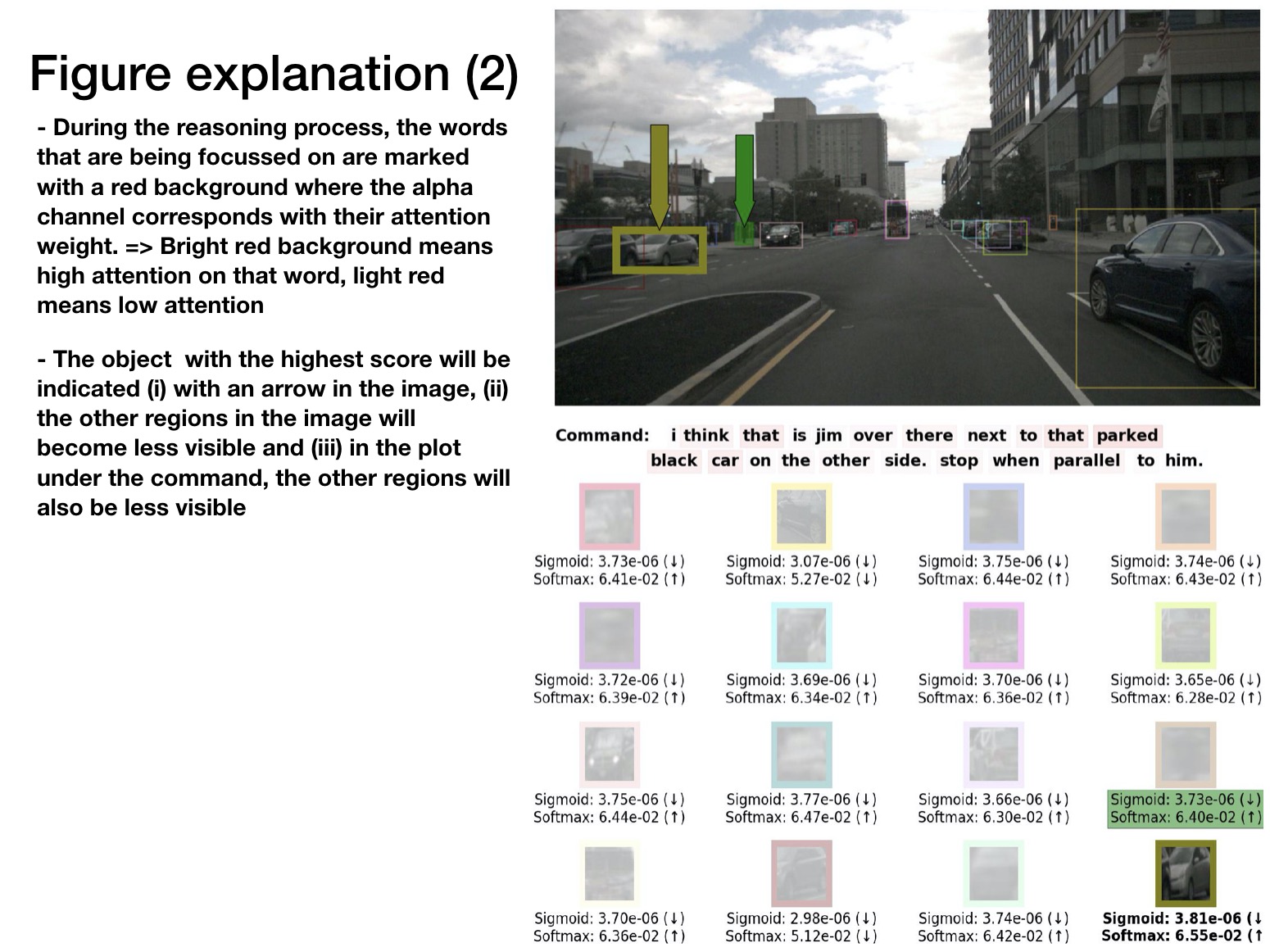}
\caption{Explaining the visualisation of the reasoning process  (Part 2). Figure from \cite{AAAI20Deruyttere}.
}
    \label{fig:vis_expl2}

\end{figure*}

\begin{figure*}[h!]
  \centering
    \includegraphics[width=\textwidth]{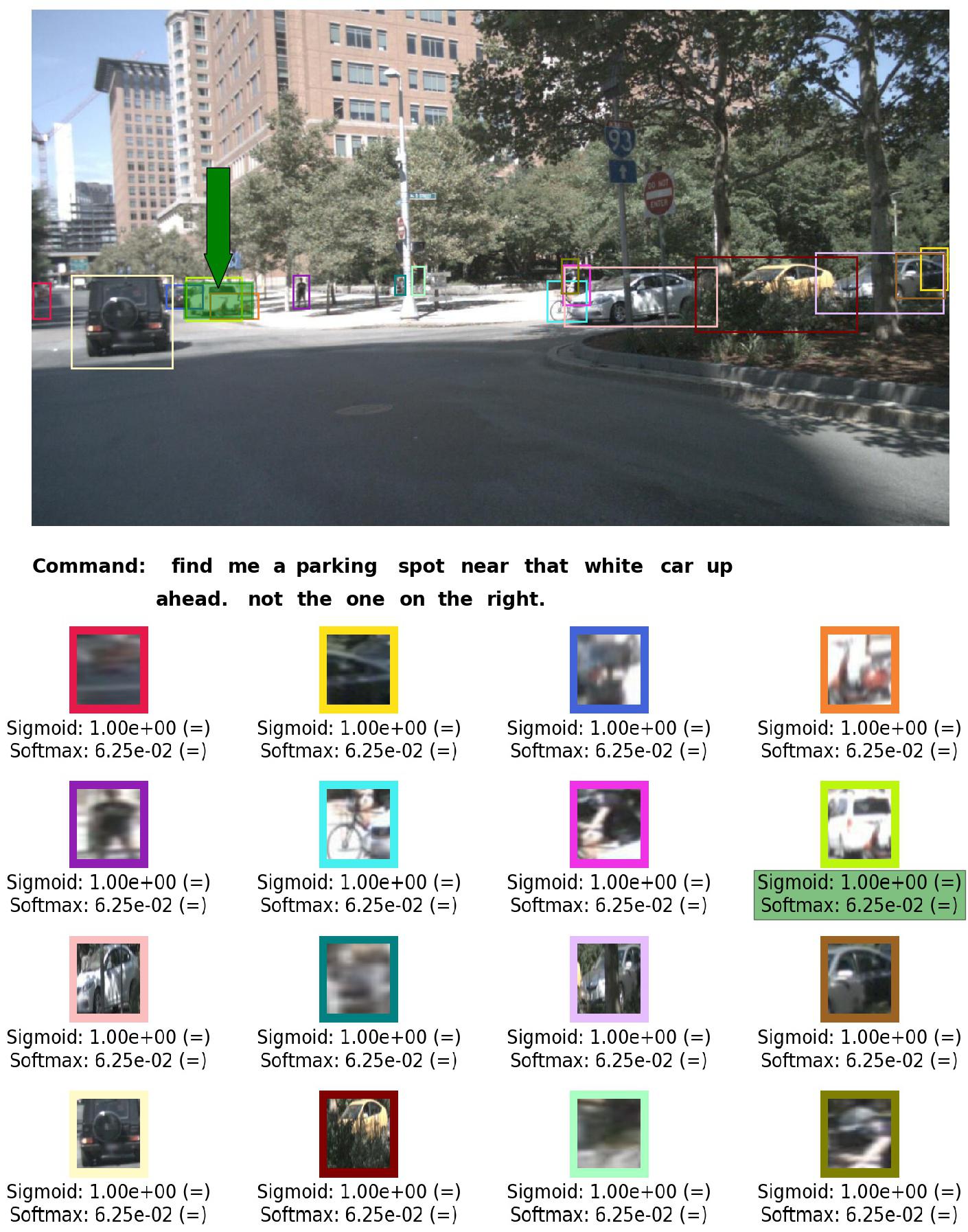}
\caption{Example 3 -  The state of the model before the reasoning process starts for the given command, regions and image. Figure from \cite{AAAI20Deruyttere}.
}
    \label{fig:enough-info0}
\end{figure*}

\begin{figure*}[h!]
  \centering
    \includegraphics[width=\textwidth]{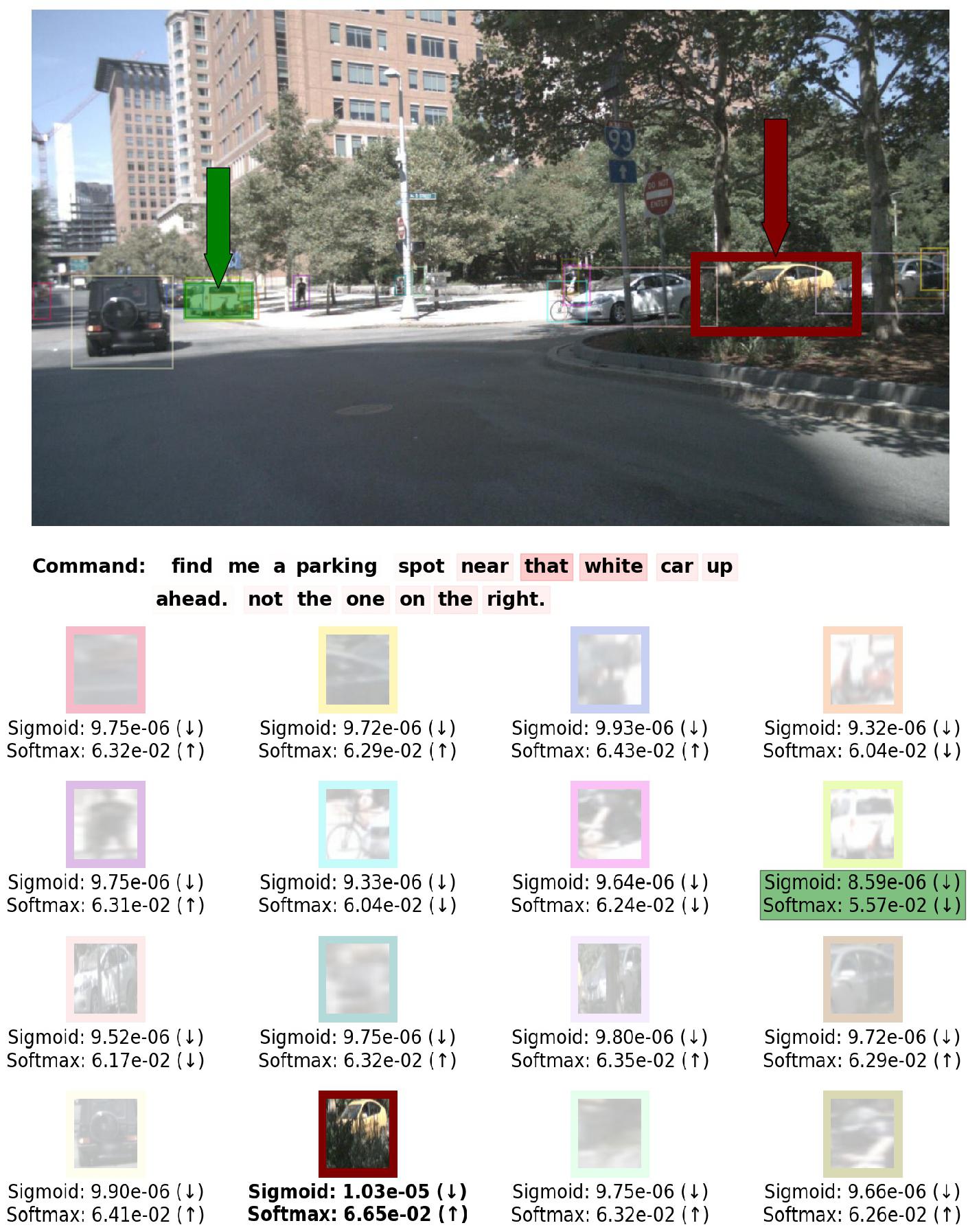}
\caption{Example 3 - Visualization of reasoning process. Step 1. Figure from \cite{AAAI20Deruyttere}.
}
    \label{fig:enough-info1}
\end{figure*}

\begin{figure*}[h!]
  \centering
    \includegraphics[width=\textwidth]{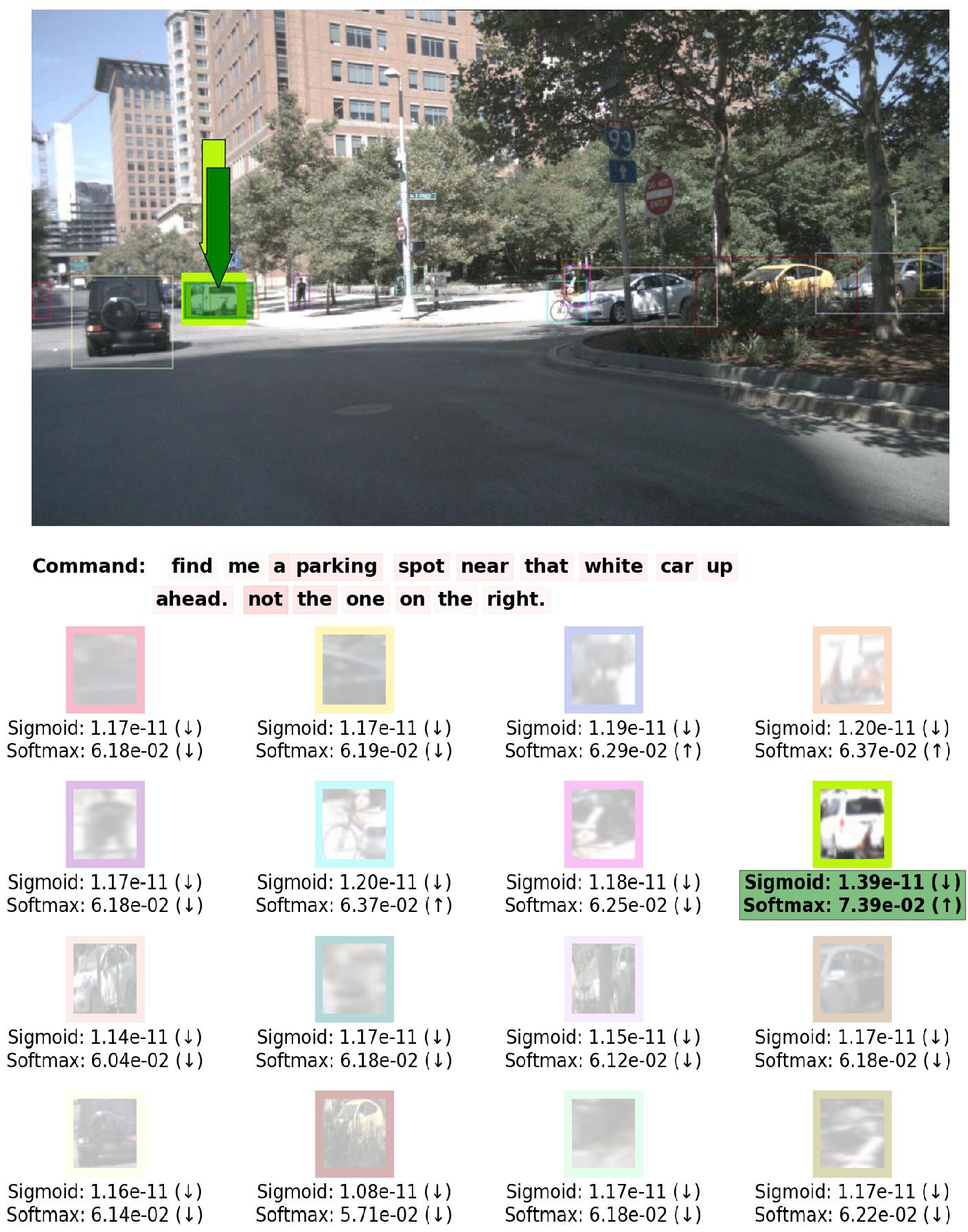}
\caption{Example 3 - Visualization of reasoning process. Step 6. Figure from \cite{AAAI20Deruyttere}.
}
    \label{fig:enough-info6}
\end{figure*}

\begin{figure*}[h!]
  \centering
    \includegraphics[width=\textwidth]{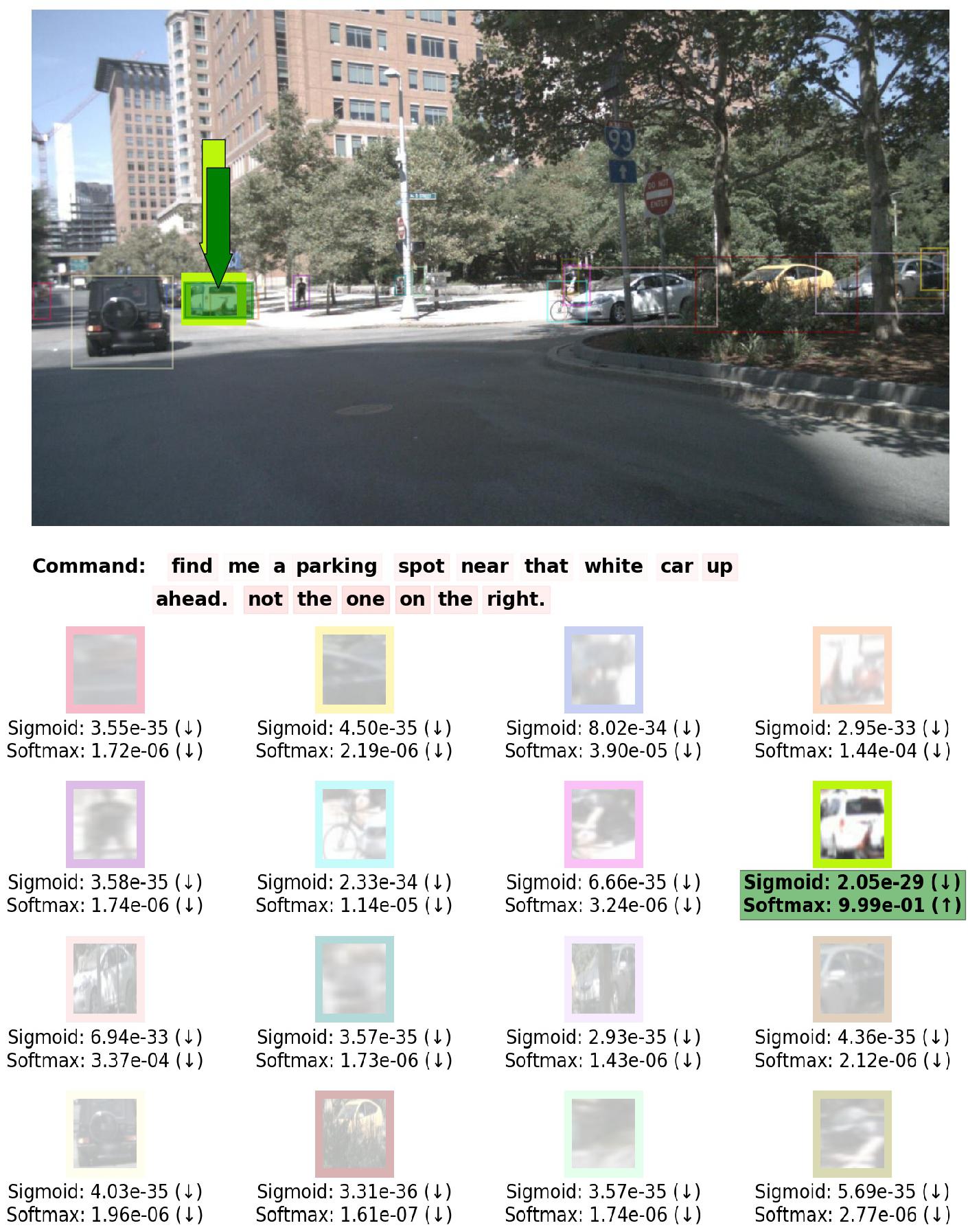}
\caption{Example 3 - Visualization of reasoning process. Final step. Figure from \cite{AAAI20Deruyttere}.
}
    \label{fig:enough-info10}
\end{figure*}

The example used in this section uses a specific visualisation that first needs to be introduced. In the Figures \ref{fig:vis_expl1} and \ref{fig:vis_expl2}, we explain in detail this visualisation. Then, Figure \ref{fig:enough-info0} shows the starting state of the MSRR. Figure \ref{fig:enough-info1} shows that the model makes a wrong decission at first but in Figure \ref{fig:enough-info6}, and after six reasoning steps, we see that the model selects the correct answer. Finally, in Figure \ref{fig:enough-info10}, we see that the object selected after six reasoning step, is the final output of the model.

% \newpage 

% \newpage

% \newpage

% \subsection{When Does a Multi-Step Model Change Decisions}
% \label{sect:MSRR_change_decision}

% In Appendix \ref{subsect:MSRR_Reasoning} we show that having multiple reasoning steps might be beneficial, we also want to show how many times such a multi-step reasoning model makes its final decision in a certain step. Assume we have a MSRR model that uses 10 reasoning steps, Figure \ref{fig:MSRR_change_decision} shows in which of these 10 reasoning steps the model makes its final prediction. It is clear that most of the final predictions are made in the very first reasoning step. For instance, if we would only consider the answers in the first step and ignore any change of decision in the following steps, we would achieve $\approx$ 55\% $IoU_{.5}$. Yet, by including more reasoning steps we can further improve this to $\approx$ 60.0 $IoU_{.5}$. This shows that having reasoning steps can be benificial for this kind of task.

% \begin{figure*}[h!]
%   \centering
%     \includegraphics[width=\textwidth]{MSRR_Final_Choice.png}
% \caption{This plot shows in which step a 10 reasoning step MSRR makes its final decision. We use MSRR Correct (blue) to indicate when the final answer by the model is also the correct answer while MSRR Wrong (orange) is used when the final answer is the wrong answer.}
%     \label{fig:MSRR_change_decision}
% \end{figure*}

\end{subappendices}

\end{document}